# Chemception:

# A Deep Neural Network with Minimal Chemistry Knowledge Matches the Performance of Expert-developed QSAR/QSPR Models

Garrett B. Goh,[*,†] Charles Siegel, Abhinav Vishnu,[†] Nathan O. Hodas,[‡] Nathan Baker[†]

[†]*High Performance Computing Group, Pacific Northwest National Laboratory, 902 Battelle Blvd, Richland, WA 99354*

[‡]*Data Science and Analytics, Pacific Northwest National Laboratory, 902 Battelle Blvd, Richland, WA 99354*

* Corresponding Author: Garrett B. Goh

Email: garrett.goh@pnnl.gov

**Abstract**

In the last few years, we have seen the transformative impact of deep learning in many applications, particularly in speech recognition and computer vision. Inspired by Google's Inception-ResNet deep convolutional neural network (CNN) for image classification, we have developed "Chemception", a deep CNN for the prediction of chemical properties, using just the images of 2D drawings of molecules. We develop Chemception without providing *any* additional explicit chemistry knowledge, such as basic concepts like periodicity, or advanced features like molecular descriptors and fingerprints. We then show how Chemception can serve as a general-purpose neural network architecture for predicting toxicity, activity, and solvation properties when trained on a modest database of 600 to 40,000 compounds. When compared to multi-layer perceptron (MLP) deep neural networks trained with ECFP fingerprints, Chemception slightly outperforms in activity and solvation prediction and slightly underperforms in toxicity prediction. Having matched the performance of expert-developed QSAR/QSPR deep learning models, our work demonstrates the plausibility of using deep neural networks to assist in computational chemistry research, where the feature engineering process is performed primarily by a deep learning algorithm.

**1.     Introduction**

ImageNet Large Scale Visual Recognition Challenge (ILSVRC) is an annual assessment and competition of various image classification computer algorithms for computer vision applications. In 2012, deep learning algorithms were first introduced to this community by Hinton and co-workers,[1] and their deep neural network (DNN) model, AlexNet, achieved a 16.4% top-5 error rate, far exceeding the 25-30% error rate for state-of-the-art models employed at that time.[1] Since then, DNN-based models have become the dominant algorithm used in computer vision, and human accuracy (less than 5% for top-5 error) was achieved by 2015, approximately 3 years after the entry of deep learning into this community.[2-3] More recently, deep learning has also begun to emerge in other fields, such as high-energy particle physics,[4-5] astrophysics,[6] and bioinformatics,[7-8] In chemistry, a few notable recent achievements include DNN-based models winning the Merck Kaggle challenge for activity prediction in 2012 and the NIH Tox21 challenge for toxicity prediction in 2014. Since then, numerous research groups have demonstrated the impact of DNN-based models to predict a wide range of properties, including activity,[9-12] toxicity,[13-14] reactivity,[15-17] solubility,[18], ADMET,[19] docking,[20] atomization energies and other quantum properties.[21-23] In recent reviews across various chemistry sub-fields, DNN-based models typically perform as well as or better than previous state-of-the-art models based on traditional machine learning algorithms such as support vector machines, and random forests.[24-25]

Unlike other machine learning algorithms, including those used by past and current computational chemistry applications, deep learning distinguishes itself in the use of a hierarchical cascade of non-linear functions. This allows it to learn representations and extract necessary features (which are conceptually similar to molecular descriptors and fingerprints in the context of chemistry) from its input data to predict the desired property of interest. This representation learning ability is the key capability that has enabled deep learning to make significant and transformative impacts in its "parent" field of computer vision. Prior to the introduction of deep learning, computer vision researchers invested substantial efforts in developing appropriate features[26]; such expert-driven development has been mostly replaced by deep learning models that automatically develop their own set of internal features, and have exceeded human-level accuracy in certain tasks.[2-3] In their current state, deep learning algorithms are not artificial general intelligence or strong "AI" systems and, as such, cannot be replace human creativity or intelligence in the scientific research process. Nevertheless, it is undeniable that deep learning have successfully demonstrated performance that is as good as, and at times superior to humans, in task-specific applications that goes beyond computer vision. For example, AlphaGo, a program based off deep neural networks outperformed top human players in Go in 2016,[27] which is a game that is over $10^{50}$ times more complex than chess, algorithms such as Google Neural Machine Translation are almost effectively as good as human translators for some languages,[28] and deep neural networks have been critical to the development of various "intelligent" software and products, including self-driving cars and personal assistant software (Siri, Cortana, etc.). In all these high impact examples that utilize deep learning algorithms, there is a common theme – what used to be in the domain of human experts, notably feature engineering, has been to a large extent replaced by the representation learning ability of deep neural networks. Such a research paradigm now drives much of modern deep learning research, as illustrated by recent works of Google's Machine Intelligence research,[29] and Microsoft AI research.[30]

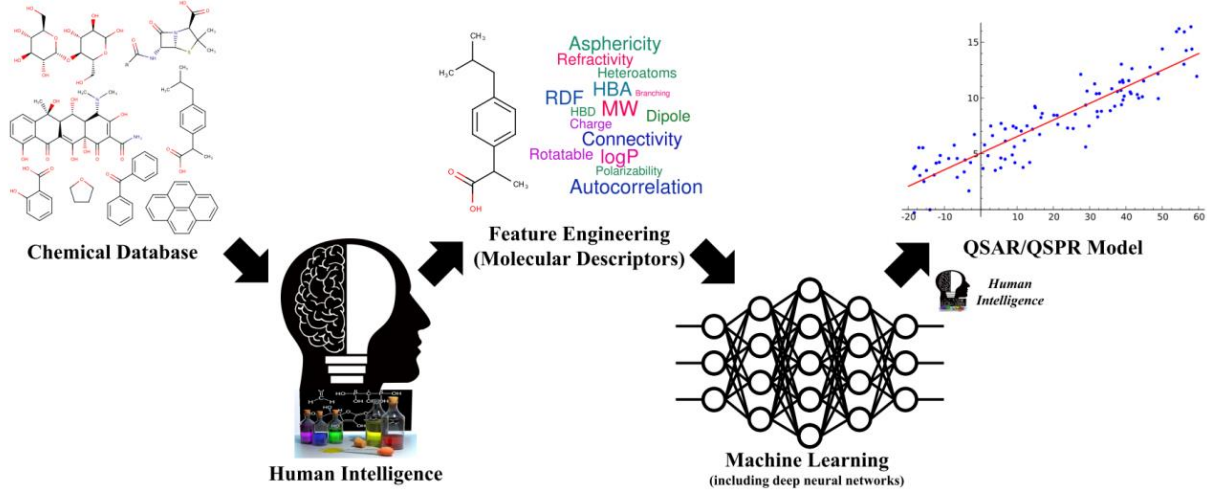

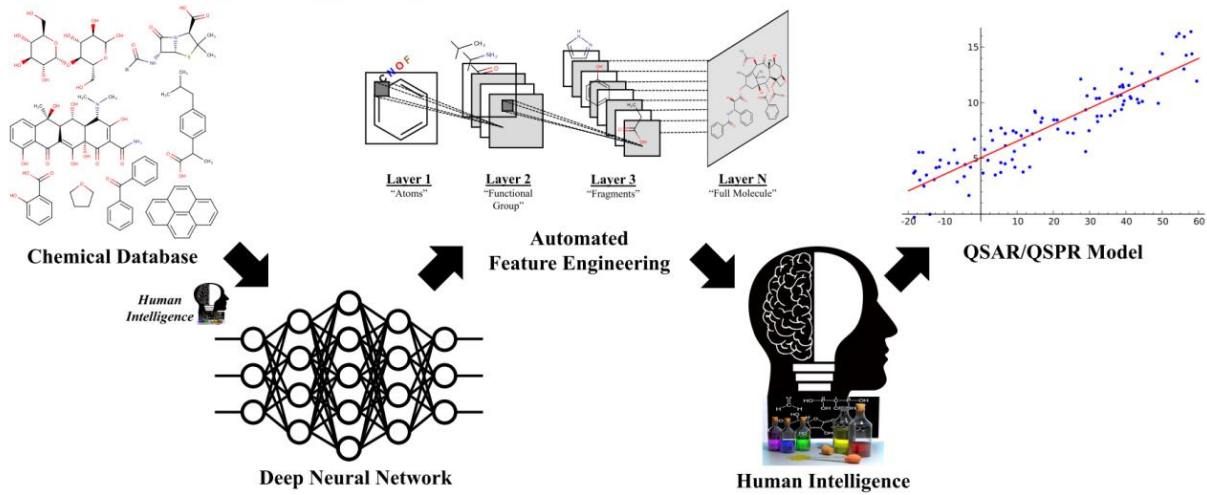

**Figure 1:** The key difference in using deep learning algorithms as a machine learning tool as opposed to a "machine intelligence" tool is the assistance, augmentation and possible replacement, for human-led tasks like feature engineering in computational chemistry.

In chemistry research, historically, human intelligence building on substantial domain-expert chemistry knowledge is used in the development of chemistry-specific features such as molecular descriptors and fingerprints. These human-expert features are then used by machine learning algorithms for QSAR and QSPR modeling, which in recent years have also included deep neural networks.[9-23] However, we observed that even in the most recent chemistry literature, deep learning in chemistry still falls under this traditional research paradigm. As illustrated in **Figure 1**, there is a subtle but important distinction that separates the approach of deep learning as a machine learning tool (as it has been traditionally done in chemistry) from the "machine intelligence" approach utilized in the above-mentioned technological advancements. We propose that a research paradigm that uses deep learning as a "machine intelligence" tool, should endeavor to assist and if possible replace, the time and resource intensive part of human-led feature engineering. In the context of our work, we utilize "raw data" in the form of 2D drawings of molecules that requires the minimal amount (i.e. no higher than high-school level) of chemical knowledge to create. We investigate the viability of augmenting and possibly eliminating human-expert feature engineering in specific computational chemistry applications. We demonstrate this by training deep convolutional neural networks to predict chemical properties that spans a broad range of categories including physical (solvation free energies), biochemical (*in vitro* HIV activity) and physiological (*in vivo* toxicity) measurements, without the input of advanced chemistry knowledge, but instead allowing the network to develop its own representations and features from the images it is trained on. This deep convolutional neural network, which we term "Chemception", despite being provided the minimal chemical knowledge, achieves a level of performance that is on par with expert-developed QSAR/QSPR models based off molecular descriptors, fingerprints and other engineered features.

The long-term significance of our work therefore lies in the amount of resources that has been expended and the speed at which Chemception achieves comparable performance for toxicity, activity and solvation predictions to current QSAR/QSPR models. In terms of computing resource, the model for the largest dataset on HIV activity can be trained within 24 hours on a single NVIDIA GTX 1080 GPU. More importantly, as Chemception was used as a "machine intelligence" tool, virtually *zero* effort was invested in developing an appropriate set of features. In contrast, in the traditional machine learning *as a tool* approach that dominates conventional QSAR/QSPR modeling, molecular descriptors and fingerprints are required inputs. Therefore, modern cheminformatics research is dependent on the substantial accumulation of chemistry research and knowledge dating back to the 1940s where the first molecular descriptors like the Wiener index and Platt number were reported in the literature.[31-32] Therefore, the development of an appropriate set of molecular descriptors and other advance features is both a non-trivial research task, and a pre-requisite for QSAR/QSPR modeling to succeed. However, as Chemception performs comparably to existing QSAR/QSPR models, its use technically obviates the necessary requirement to develop appropriate molecular descriptors and fingerprints for cheminformatics research, which will be useful in situations where the current generation of expert-developed features fail to provide sufficient performance. In addition, the fact that the same Chemception architecture performs similarly well for different chemical properties, without the need for tuning the network design for each property, suggests that Chemception is a general-purpose neural network that can be used to predict other properties of small-molecules beyond those studied in this paper. Lastly, as we have taken a domain agnostic approach in developing Chemception, we anticipate that the general approach used will also be transferable to other types of research challenges, particularly if an appropriate pictorial representation of the data exist.

## 2. Methods

### 2.1 Computational Framework Overview

As illustrated in **Figure 2**, the end-to-end workflow of the Chemception framework requires minimal chemical knowledge beyond the generation of 2D chemical structures. SMILES strings were converted to their respective 2D molecular structures, which were then mapped onto an input array used to train the convolutional neural network in a supervised fashion. Apart from the generation of chemical images and the measured chemical properties used to train the model; no additional source of chemistry-inspired features, such as molecular descriptors or fingerprints were used.

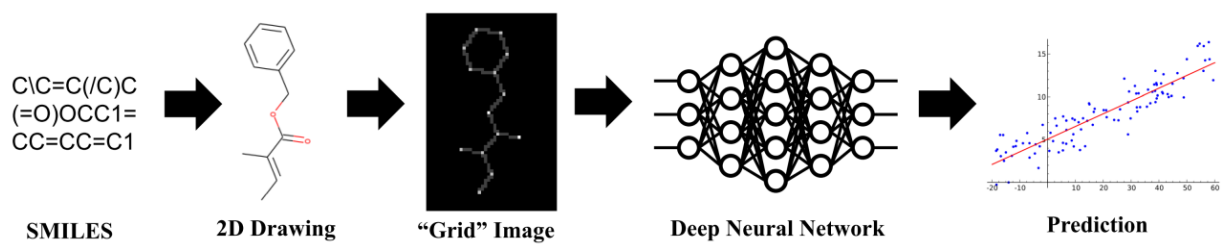

**Figure 2:** Illustration of the Chemception framework. After a SMILES to structure conversion, the 2D images are mapped onto an 80 x 80 image that serves as the input image data for training a deep neural network to predict toxicity, activity, and solvation properties.

### 2.2 Data Preparation

Data preparation was performed using Python, utilizing script bindings to open-source cheminformatics software, including OpenBabel[33], Pybel[34] and RDKit[35] through the Cinfony interface.[36] Chemicals in the databases are stored as SMILES strings,[37-38] which are compact string representation that describes a molecule's structure. The SMILES string were decoded to their corresponding 2D molecular structure using the above-mentioned cheminformatics software. The resulting coordinates of the 2D structure of each molecule were then mapped onto a 80 x 80 grid, where each pixel had a resolution of 0.5 Å as illustrated in **Figure 2**. The resulting 80 x 80 array was then greyscale "color coded" based on the presence of an atom or a bond. Specifically, atoms mapped onto the grid were assigned a number based on its atomic mass unit, and bonds mapped onto the grid were assigned the number 2, as it does not correspond to the identity of any element in the training set. The other parts of the grid were empty (i.e. vacuum) and were defaulted to the number 0. The resulting discretized image of the molecule was then parsed into a deep convolutional neural network for training.

## 2.3    Chemistry Knowledge Used

While open-source cheminformatics software was used in the data preparation stage, its role was to automate the data preparation process and facilitate compatibility with existing chemical databases. *In other words, using a database of photographs of consistently and correctly drawn chemical structures that requires no more than high school level chemistry knowledge as input data, would be equivalent to using the computer-generated structures generated by the data preparation workflow as mentioned above.* In addition, it should be emphasized that no additional source of chemistry-inspired features or inputs, such as molecular descriptors or fingerprints were used in training the model. This means that Chemception was not explicitly provided with even the most basic chemical concepts like "valency" or "periodicity". Given our choice to encode the image in a single-channel greyscale where each atom and bond has a unique number, there is the *potential* for the neural network to learn basic chemistry concepts if that would facilitate the prediction of the property of interest. However, such a task may not be particularly straightforward because atom information, bond information, and information about empty space (vacuum) all share the same channel in the image. Therefore, the prediction of the molecule's properties is entirely dependent on Chemception's ability to learn relevant representations from the image data and automatically engineer its own features (i.e. its own analogous set of "molecular descriptors") relevant for predicting the chemical property of interest.

## 2.4 Dataset Description

**Table 1:** Summary of datasets used to evaluate the performance of Chemception

| Dataset | Property | Task | Size |
|---------|----------|------|------|
| Tox21 | Physiological: Toxicity | Multi-task binary classification | 8014 |
| HIV | Biochemical: Activity | Single-task binary classification | 41,193 |
| FreeSolv | Physical: Free energy of solvation | Single-task regression | 643 |

We obtained several publically available datasets (**Table 1**), mostly from the MoleculeNet benchmark database,[39] to evaluate the performance of Chemception. The Tox21 dataset[40] is a NIH-funded public database of toxicity measurements comprising of 8014 compounds on 12 different measurements ranging from stress response pathways to nuclear receptors. This dataset provides a binary classification problem of labeling molecules as either "toxic" or "non-toxic". In addition to Tox21 that measures physiological effects of chemicals, we also examined two other datasets that focus on different type of properties – physical and biochemical. We evaluated the performance of Chemception on the FreeSolv dataset, which comprises 643 compounds that have measured hydration free energies of small-molecules ranging from –25.5 to 3.4 kcal/mol.[41] Hydration free energy is a physical property of the molecule which can be computed from first principles.[41] The dataset also included alchemical free energy calculations obtained from molecular dynamics simulations, and these were used as target comparisons for physics-based models. Lastly, we evaluated the performance of Chemception on the HIV dataset obtained from the Drug Therapeutics Program AIDS Antiviral Screen, which measured the ability of 41,913 compounds to inhibit HIV replication *in vitro*.[42] Using the curation methodology adopted by MoleculeNet,[39] this dataset was reduced to a binary classification problem of "active" and "inactive" compounds.

**2.5     Dataset Preprocessing**

In processing the Chemception training data, we used a 5-fold cross validation protocol for training, and evaluated the performance and early stopping criterion of the model using the validation set. We also included the performance on a separate test set as an indicator of generalizability. Specifically, for the Tox21 and HIV dataset, $1/6^{th}$ of the database was separated out to form the test set, and for the Freesolv dataset, owing to its smaller size, $1/10^{th}$ of the database was used to form the test set. The remaining $5/6^{th}$ or $9/10^{th}$ of the dataset was then used in the random 5-fold cross validation approach for training Chemception, and stratification was enforced for the classification tasks (Tox21, HIV), which meant that the training, validation and test set had approximately the same ratio of classes. In addition, we noted that the ratio of the classes for classification tasks were imbalanced, ranging from 1:5 to 1:34. To address this skewed distribution, we oversampled the minority class; in the construction of the training/validation/test set, we computed the imbalance ratio and appended additional data from the minority class by that imbalance ratio. It should be noted that this oversampling step was performed after stratification – which means that while the same data may be repeated in a particular training/validation/test set, the data across different training/validation/test sets are unique and do not overlap.

**2.6     Deep Neural Networks**

The theory of deep neural networks have been extensively documented in prior publications and reviews.[43-45] For the purpose of this manuscript, in this section, we briefly introduce the high-level conceptual details that are necessary for understanding deep neural networks. Artificial neural networks, on which deep learning algorithms are based on, are a class of machine learning algorithms used to model and analyze large complex datasets. The basic unit of neural networks is a "neuron", and for conceptual and computational convenience, these neurons are organized into layers, whose network design is inspired by biological neural networks. Each neuron in the network performs a computation that converts the input data received into an output value, by mapping it onto a non-linear function. In addition, a tunable parameter, the "weight" of each neuron's function is adjusted in the construction of the model to minimize the error of the predicted value, a process known as "training" the neural network. Operationally, the input is represented as a vector and the weights of the neurons in the layer are arranged into a matrix. This layer then performs matrix-vector multiplication followed by a nonlinear activation function. For most modern neural networks, including the Chemception neural network developed in this paper, rectified linear activation functions (ReLU)[46] are used, as this specific functional form enables the training of many layers to form a deep neural network.

During training, it is necessary to determine how to assign error attribution and make corrections to its weights by working backwards originating from the predicted output, and back through the neural network. This backwards propagation of errors is known formally as "backpropagation". During backpropagation, the gradient descent algorithm is used to find the minimum in the error surface caused by each respective neuron when generating a corresponding output. Conceptually, gradient descent is no different from the steepest descent algorithm used in classical molecular dynamics simulation. The key difference is instead of iteratively minimizing an energy function and updating atomic coordinates for each step, a loss function of the target output of the DNN is iteratively minimized and the weights of the neurons are updated each step, which are also known as "iteration" in the literature. The data in the training set may be iterated over multiple times, with a complete pass over the data being called an "epoch."

Convolutional Neural Networks (CNN) are a special type of neural network developed to handle image data, extract and engineer features relevant for the task it is trained on. Since 2012, every winning entry in the annual ImageNet competition has been based primarily on a CNN architecture, although additional tweaks and variations in architecture design have emerged over the years.[29-30] As illustrated in **Figure 3a**, a CNN is constructed from convolutional layers; instead of having every neuron in each layer connected to every neuron in the previous layer, each neuron in a convolutional layer (or "filter" as it is more commonly termed in the literature) only receives input from a small, spatially contiguous window on the output of the previous layer. Furthermore, there are filters that receive windows across the entire input with shared weights, so that they all detect the same feature in different locations in an image. This allows convolutional layers to preserve spatial structure in data, such as the relative positions of pixels in images. In setting up convolution layers, it is necessary to specify the spatially contiguous window on which each filter operates on. Formally, this is referred as the size; a 4x4 convolutional layer thus operates on a spatial region that is 4x4 "pixels" in dimensions. In addition, the "overlap" between adjacent windows is governed by the stride; a 4x4 convolution layer with a stride of 2 would thus downsample the image into regions of 4x4 "pixels", and each region would have a 2 "pixel" overlap with one another.

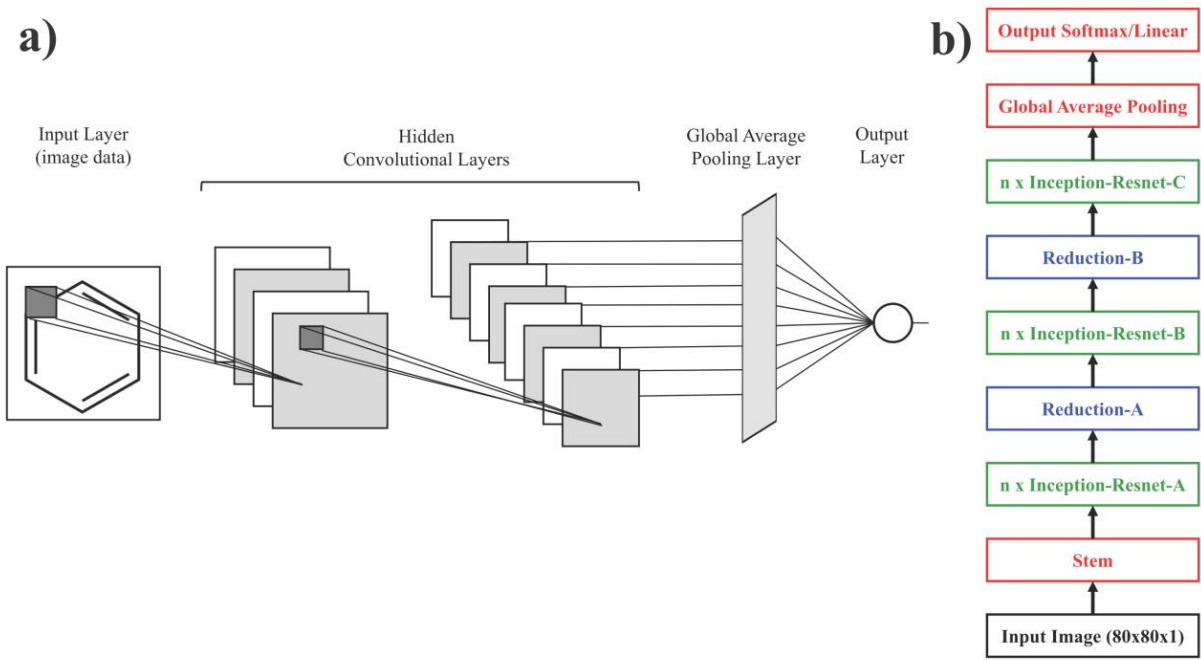

**Figure 3:** (a) Depiction of a typical convolutional neural network, and (b) High-level architectural details of the Chemception architecture. Each colored box denote a different segment: the stem segment has a single 4x4 convolutional layer, with stride 2, the Inception-Resnet segments have varying (1 to 3) number of inception blocks stacked sequentially, and the reduction segments have a single block. Architectural details of the inception/reduction blocks are elaborated in Figure 4.

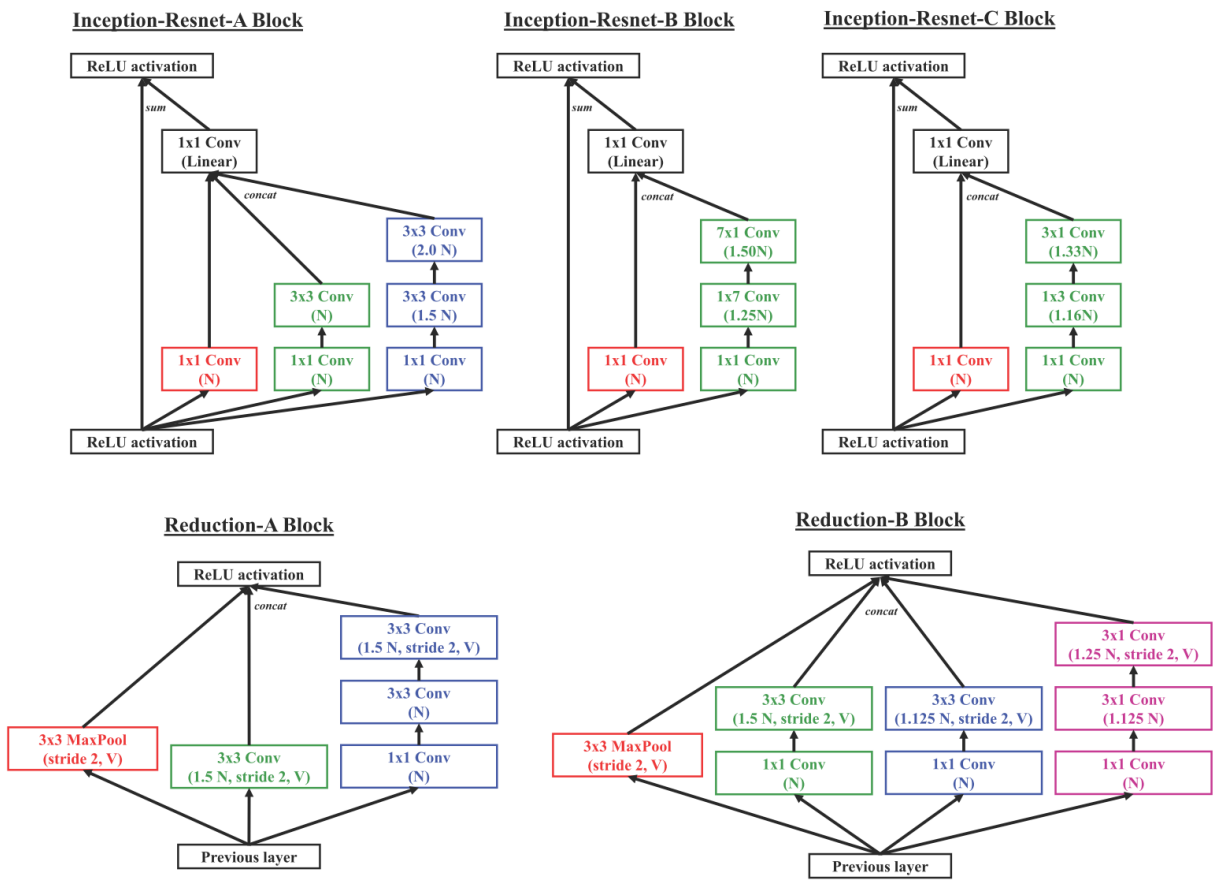

**Figure 4:** Architectural details of the various inception/reduction blocks in each segment of Chemception. Layers are denoted in colored boxes, and are assumed to have a ReLU activation layer after the specified convolution layer, with a stride of 1, and "same" padding unless otherwise noted. Each block has N convolutional filters for each layer, and the variations are indicated as multiples of N. Exceptions include: inception blocks concatenate to a final layer that has a linear activation (instead of ReLU), and all final layers of each branch of the reduction block have a stride of 2 and "valid" padding.

## 2.7 Convolutional Neural Network Design

For our work, we developed Chemception based on the Inception-ResNet v2 neural network architecture,[47] that combines arguably the two most important architecture advances in CNN design since the debut of AlexNet in 2012 - Inception modules[29] and deep residual learning.[30] Similar to Inception-ResNet v2, Chemception high level architectural design includes 6 segments. As illustrated in **Figure 3b**, this includes the stem layer, followed by a series of Inception-ResNet A blocks, a reduction block (Reduction-A), a second series of Inception-ResNet B blocks, a second reduction block (Reduction-B), the third and last series of Inception-ResNet C blocks, which is then passed to a global pooling layer that leads directly to the final output layer. For classification problems (Tox21, HIV), a softmax layer was used as the output layer, and for regression problems (FreeSolv), a linear layer was used as the output layer.

In the development of Chemception, and given that that Inception-ResNet v2 is a highly optimized architecture for processing image data, we did not modify many architectural regularities of the neural network, which includes the design choices of the various inception modules and reduction modules. We did however, optimize the number of inception blocks in each segment, and also the width of the layers in each block. We observed that in the original Inception-ResNet v2 design, within each inception block, there was a specific regularity in the number of convolutional filters across various branches in the inception module, and also within each branch itself, and this regular pattern was retained. As illustrated in **Figure 4**, the inception and reduction blocks from Inception-ResNet v2 were converted to a reference design block. This reference design had a selected reference layer for each block, from which the convolutional filters of other layers are calculated, while maintaining the observed regularity of the block design. Subsequently for the rest of this paper, when we indicate that the number of convolutional filters for a particular block is set to a specific number, we are referring to the number of convolutional filters of the reference layer for that particular block. Lastly, instead of using a series of 7 layers for the stem segment as in Inception-ResNet v2, we replaced it with a single convolutional layer of size (4, 4) and stride (2, 2).

A baseline model, Chemception T1, which represents the shortest possible design was constructed from these high level architectural details. We also tested an additional 9 iterations of the Chemception architecture, and the design details and performance are included with the results.

## 2.8 Chemception Training Protocol

Chemception was trained using a standard 5-fold cross validation protocol, and the details of the dataset splitting have been included in the preceding section. Chemception was trained using a tensorflow backend[48] with GPU acceleration using NVIDIA CuDNN libraries.[49] The network was created and executed using the Keras 1.2 functional API interface.[50] Chemception was trained using a two-stage protocol. In the first stage, we use the RMSprop algorithm[51] for 50 epochs using the standard settings recommended (learning rate = $10^{-3}$, $\rho = 0.9$, $\varepsilon = 10^{-8}$). This was followed by a second fine-tuning stage using the stochastic gradient descent (SGD) algorithm with momentum for another 50 epochs, using an initial learning rate of $10^{-3}$ with an exponential learning rate decay mapped using the following function:

$$lr = lr_{ini} * \gamma^{epoch}$$

where lr denotes the current learning rate, $lr_{ini}$ denotes the initial learning rate, $\gamma$ is an empirical scaling factor that is set to 0.92, and epoch denotes the current epoch. We used a batch size of 32, and after class balancing and oversampling, each task in the Tox21 dataset was trained (refers to combined size of training and validation set) on ~8,000 to ~12,000 compounds. The HIV dataset was trained on ~66,000 compounds and Freesolv was trained on ~580 compounds. For a full 100 epoch training cycle, this means that Chemception would observe on average ~30,000 batch updates. For the Freesolv dataset, because of its small size, we also added 10X more data per epoch to ensure it had the same order of magnitude of batch updates as the Tox21 and HIV datasets. Lastly, with a large number of trainable parameters in Chemception compared to traditional machine learning algorithms, which ranges between ~70,000 to ~2,300,000 depending on the Chemception architecture used, we also included an early stopping protocol to reduce overfitting. This was done by monitoring the loss of the validation set, and if there was no improvement in the validation loss after 25 epochs, the last best model as evaluated by the validation loss was saved, and this was used for the second fine tuning stage of training or saved as the final model.

Finally, for the Tox21 and HIV dataset, the evaluation metric reported in our paper is area under the ROC-curve (AUC). For the FreeSolv dataset, the evaluation metric is RMSE, which for a given dataset of *n* samples is defined as:

$$\text{RMSE} = \sqrt{\frac{1}{n}\sum_{i=1}^{n}(x_{i,predicted} - x_{i,measured})^2}$$

The reported results in the paper are the mean value and standard deviation of the performance metric, obtained from the 5 runs in the 5-fold cross validation protocol.

## 2.8  Data Augmentation on Input Images

During the training of Chemception, we performed additional real-time data augmentation to the image using the ImageDataGenerator function in the Keras API, so as to bolster the limited number of data that we have for each task. Such data augmentation techniques are a common and recommended practice in the computer vision literature[52], and include operations such as shearing, cropping, rotation, etc. However, unlike photographic images where a substantial part of the image is used in making the prediction,[53] we hypothesize this may not be the case for our data. This is because the images of the molecules are comparatively sparser as they are populated mostly (>90%) by zeros, and the "usable" data is thus localized to a very small fraction of the image. Paradoxically, while the image may be sparse, each pixel is also richer in terms of information density. To illustrate this point, a hydroxyl (-OH) group requires a minimum of 2 pixels to draw – one pixel for the atom, and another pixel for the bond. For the hydroxyl group to have some background molecular context, the relevant information may include bond pixels to the adjacent atom pixels, and as little as ~10 pixels may therefore be sufficient to describe a hydroxyl group and its local chemical environment. In the context of an 80 x 80 image, 10 pixels represents only 0.156% of the entire image area. Consequently, data augmentation techniques that distort the fidelity of the data, such as random cropping was not used, which in this example may lead to a loss of part of the molecule. In our data augmentation protocol, each image was randomly rotated between 0 to 180 degrees before being parsed into Chemception. Incidentally, it should be noted that this procedure also facilitates Chemception to learn rotational invariance of the chemical structures it sees.

## 3. Results and Discussion

### 3.1 The Chemception Architecture

The pictorial format of our data suggests that deep convolutional neural networks architecture and designs developed in the computer vision domain could be adopted for chemical research purposes. Here, we detail our design and justification for a novel "Chemception" neural network architecture that is based on Inception-ResNet v2.[47]

In the post AlexNet-era (from 2012 to present), the drive to achieve human-level performance of under 5% top-5 error rate in ImageNet competition, was driven by better architectures rather than more data. Notably, Inception modules developed by Google allowed GoogleNet to attain a 6.67% top-5 error rate and it was the winning entry for the ImageNet competition in 2014. The key conceptual advance of Inception modules, is that instead of viewing the image at different spatial levels and attempting to draw correlations like in a traditional CNN design, the Inception module "forces" the network to look at the data at different spatial levels at the same time by concatenating the data (**Figure 5a**). We hypothesize that this type of architecture would be useful, as it encourages correlation and development of representations that link various spatial levels, which in the chemistry context could include forming representations that link the atomic level to the functional group level to the fragment level and ultimately to the whole molecule level. In 2015, Microsoft developed ResNet, and its key invention, residual training, allowed for the training of much deeper neural networks, up to 1000 layers if necessary,[30] and ResNet-based models were the winning entries in ImageNet competitions in both 2015 and 2016. A residual link (**Figure 5b**) basically provides the option for the network to "skip" the layer by using an identity function if no usable information is being learned. As residual links sums the last layer's output to the next layer's output, it also means that the network is now training a function near the identity rather than one near zero, which makes it practically easier to train neural networks of excessive depths (>50 layers). In 2016, Google reported a newer architecture combining Inception modules with residual links.(**Figure 5b**) Specifically, Inception-ResNet v2 was the template model used for the development of our Chemception architecture.

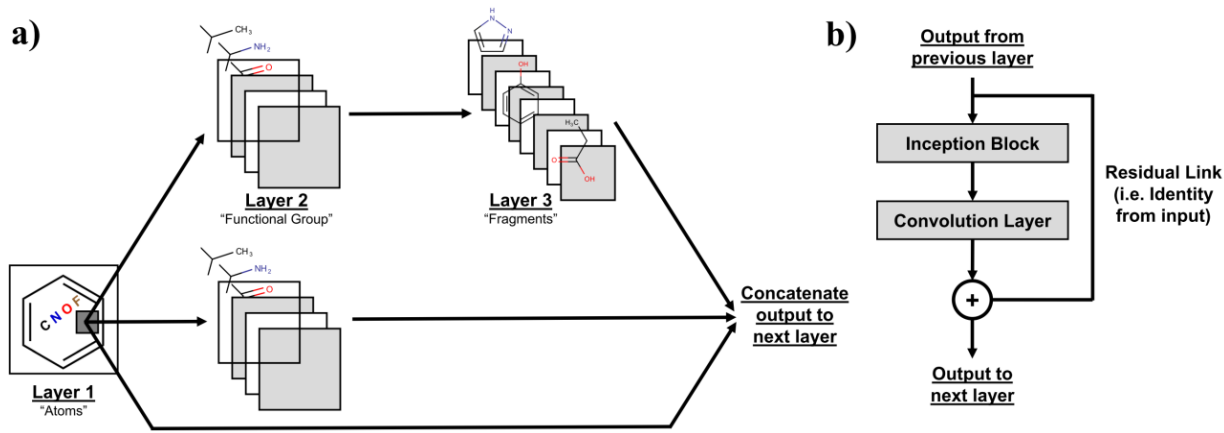

**Figure 5:** (a) Illustration of the Inception module, which concatenates data at different spatial levels. (b) Illustration of a residual link as used in Inception-ResNet v2.

The original Inception-ResNet v2 was developed for a significantly larger dataset of 1.3 million images for a 1000-way classification problem. Hence, we hypothesized that it had a superfluous and redundant number of parameters for processing the sparser images of chemical structures in a binary classification (for Tox21, HIV) or a regression (for FreeSolv) task, on a much smaller dataset of ~600 to ~40,000 chemical images. To test this hypothesis, we started with a baseline Chemception architecture (**Figure 3b**), that comprises one block in each major segment of the Inception-ResNet v2 architecture. This model alone itself is 21 layers deep. In contrast, the original Inception-ResNet v2 had 5/10/5 inception blocks respectively and is 95 layers deep.[47] In addition, we reduced the number of convolutional filters to 32 per block, but retained most of the architectural design choices of the inception and reduction blocks (refer to the methods section for architectural details). Lastly, for simplicity, we also kept the number of filters across blocks identical, resulting in an approximately "fixed width" architecture. In contrast, the original Inception-ResNet v2 had a varying number of convolution filters, from 64 to 388.[47] However, as both the type of images and quantity of data used in our work differs substantially from typical computer vision research, we therefore decided on using the fixed width architecture as an "Occam's Razor" baseline model.

### 3.2 Chemception Model Exploration

With the establishment of the baseline Chemception architecture, hereon referred to Chemception T1 (Tier 1), we also investigated additional modifications to the baseline model. Specifically, we expanded the number of inception blocks from 1 to 2 and 3, which results in Chemception T2 and T3 respectively. In addition, we also explored uniformly shrinking and expanding the width of the layers to 16 and 64 filters respectively. A list of the various architectural iterations and the number of trainable parameters is summarized in **Table 2**. Considering the numerous architecture and tasks investigated in this paper, we also developed a streamlined training protocol to economize compute resource usage. As the database of chemical images we have generated are sparse and have little known counterparts in computer vision research, the optimized learning rate hyperparameters on natural images would not be transferable. To eschew hyperparameter optimization of learning rate schedule, we developed a standardized two-stage learning protocol. First, we use RMSprop,[51] an algorithm that automatically adjusts the learning parameters in response to the gradients, which is useful when no *a priori* information is known about the dataset. This is followed by a second fine-tuning stage using SGD with momentum, with an exponential decay learning rate to ensure convergence. Lastly, overfitting was mitigated primarily using standard data augmentation techniques and an early stopping criteria, and the detailed training protocol is elaborated in the methods section.

**Table 2:** High level summary of various iterations of Chemception architecture explored.

| Model | No. of Inception Block per segment | No. of conv filters per block | No. of parameters |
|---|---|---|---|
| Chemception T1 | 1 | 32 | 276,603 |
| Chemception T1_F16 | 1 | 16 | 69,875 |
| Chemception T1_F64 | 1 | 64 | 1,100,967 |
| Chemception T2 | 2 | 32 | 435,516 |
| Chemception T2_F16 | 2 | 16 | 109,808 |
| Chemception T2_F64 | 2 | 64 | 1,735,324 |
| Chemception T3 | 3 | 32 | 594,429 |
| Chemception T3_F16 | 3 | 16 | 149,741 |
| Chemception T3_F64 | 3 | 64 | 2,369,681 |

**3.3 Benchmark Database Selection and Controlling Performance Factors**

Evaluating different machine learning (and deep learning) models in a comparable manner is a difficult task. This is because there are multiple factors that can affect the performance of a model, and major factors include: (i) quantity of data, (ii) quality of data, (iii) representation of data and/or features, (iv) algorithm used, and (v) hyperparameter settings. In this work, the focus of our research is to determine if a deep neural network can reasonably replace human intelligence in feature engineering – the development of chemistry-relevant features such as molecular descriptors, which is also factor (iii) listed above. It is also well known that for many machine learning application, more data frequently trumps better algorithms.[54-55] Standardized competitions such as ImageNet have become the preferred option for computer vision researchers, which are reasonably robust as the data used is identical across all teams. In chemistry, the Kaggle competition by Merck released only molecular descriptors and thus would not serve as an appropriate dataset. Therefore, we have elected to use datasets from MoleculeNet,[39] and their reported results, as the best appropriate comparison to the results of Chemception, as it controls for both quantity and quality of data. Specifically, we compare our results to multilayer perceptron (MLP) deep neural networks trained on ECFP fingerprint, which is analogous to expert-developed QSAR/QSPR deep learning models reported in the literature.

We selected the Tox21 dataset as the main dataset for our subsequent study of architecture optimization, as it includes 12 different toxicity measurements for almost 10,000 compounds, and this will minimize the likelihood that good performance in a particular architecture is an anomalous result. Furthermore, to demonstrate the generalizability of Chemception, we also predicted additional properties from the physical domain (solvation energies, ~600 compounds) and the biochemical domain (HIV activity, ~40,000 compounds). In evaluating both Tox21 and HIV datasets, which are binary classification problems of active/toxic compounds, we observed that the distribution is skewed towards the non-active/toxic. In particular Tox21 is imbalanced by as much as 1:34 and HIV is imbalanced by 1:27. Such an imbalance can be detrimental to the training of the neural network. Therefore, we leveraged established protocols in machine learning, such as oversampling the minority class while maintaining proper stratification during cross-validation. In addition, we also adopted standard data augmentation techniques used in computer vision when processing the images. These data augmentation technique increases the overall size of the training set, without the use of additional labeled data, and consequently do not change the data requirement concerns as listed earlier. Further details on data sampling and augmentation are included in the methods section.

### 3.4 Model Optimization Results

We trained the baseline Chemception T1 network on the Tox21 dataset and the results are summarized in **Table 3**. The individual toxicity measurements in the Tox21 dataset were predicted with validation AUC that ranged from 0.702 to 0.834, with the mean AUC value for the entire

Tox21 dataset at 0.760. The standard deviations of each AUC value across the 5-fold cross validation runs are also reported, and we observed that 10 out of the 12 toxicity measurements had a test AUC and was within one standard deviation of the corresponding validation AUC, and all measurements were within two standard deviations of the validation AUC. This indicates that our training protocol was robust and prevented overfitting despite the substantial number of 276,603 learnable parameters in the Chemception T1 network.

**Table 3:** Summary of Results for Tox21 trained on Chemception T1 network.

|  | **Train AUC** | | | **Validation AUC** | | | **Test AUC** | | | |
|---|---|---|---|---|---|---|---|---|---|---|
| nr-ahr | 0.825 | +/- | 0.018 | 0.779 | +/- | 0.015 | 0.800 | +/- | 0.020 | Y |
| nr-ar | 0.843 | +/- | 0.010 | 0.797 | +/- | 0.049 | 0.757 | +/- | 0.029 | Y |
| nr-ar-lbd | 0.887 | +/- | 0.034 | 0.834 | +/- | 0.046 | 0.886 | +/- | 0.014 | Y |
| nr-aromatase | 0.801 | +/- | 0.010 | 0.759 | +/- | 0.027 | 0.799 | +/- | 0.016 | Y |
| nr-er | 0.747 | +/- | 0.020 | 0.710 | +/- | 0.023 | 0.694 | +/- | 0.013 | Y |
| nr-er-lbd | 0.824 | +/- | 0.029 | 0.765 | +/- | 0.036 | 0.762 | +/- | 0.009 | Y |
| nr-ppar-gamma | 0.791 | +/- | 0.038 | 0.742 | +/- | 0.025 | 0.819 | +/- | 0.015 | Y |
| sr-are | 0.724 | +/- | 0.009 | 0.702 | +/- | 0.025 | 0.654 | +/- | 0.009 | N |
| sr-atad55 | 0.841 | +/- | 0.022 | 0.759 | +/- | 0.048 | 0.776 | +/- | 0.011 | Y |
| sr-hse | 0.776 | +/- | 0.032 | 0.732 | +/- | 0.013 | 0.717 | +/- | 0.018 | N |
| sr-mmp | 0.791 | +/- | 0.020 | 0.759 | +/- | 0.016 | 0.755 | +/- | 0.010 | Y |
| sr-p53 | 0.844 | +/- | 0.034 | 0.782 | +/- | 0.036 | 0.776 | +/- | 0.011 | Y |
| **Tox21** | **0.808** | | **0.044** | **0.760** | | **0.035** | **0.766** | | **0.058** | |

Next, we expanded on the baseline network by testing 9 related iterations based on the Chemception architecture of varying depths and widths (see **Table 2** for architectural details). The results are summarized in **Table 4**, and for brevity we only included the mean AUC metrics across all 12 measurements in the Tox21 database. Our results indicate that for the shorter network depth (T1, T2), increasing the width of the layers provide no statistically significant improvement in the overall performance as measured by validation AUC. The deepest network design (T3), with a skinnier network topology with 16 convolutional filters at each block's reference layer, provided a small boost in validation AUC, from 0.760 to 0.768. Conversely, a fatter network topology with 64 convolutional filters suffered in terms of performance as the validation AUC dropped precipitously to 0.733. These observations indicate that for our dataset, a deeper and skinnier Chemception architecture might be advantageous, which is in contrast to traditional computer vision architecture designs like Inception-ResNet v2, where no one layer has as little as 16 convolutional filters.

Given the sparsity of chemical images, we hypothesize that the number of features "theoretically" learnable at each spatial level would arguably be less than a typical natural image photograph used in computer vision research. We also know that within the image of a molecule, is encoded complex chemical concepts like toxicity, hence a deeper network with more expressive power is anticipated to provide additional benefit to its learning and performance. From this perspective, it is unsurprising that a skinnier and deeper network would be a more optimal architectural choice. Next, we performed subsequent experiments of increasingly deeper Chemception architectures, and fixing the number of convolutional filters at 16. As shown in **Table S1**, the best performance plateaued at the Chemception T3 architecture with a validation AUC of 0.768 for the Tox21 dataset.

**Table 4:** Summary of Results for Tox21 trained on Chemception trial architectures

| Architecture | Train AUC | | | Validation AUC | | | Test AUC | | |
|---|---|---|---|---|---|---|---|---|---|
| Chemception_T1_F16 | 0.810 | +/- | 0.035 | 0.758 | +/- | 0.033 | 0.766 | +/- | 0.051 |
| Chemception_T1_F32 | 0.808 | +/- | 0.044 | 0.760 | +/- | 0.035 | 0.766 | +/- | 0.058 |
| Chemception_T1_F64 | 0.805 | +/- | 0.043 | 0.758 | +/- | 0.034 | 0.765 | +/- | 0.055 |
| Chemception_T2_F16 | 0.805 | +/- | 0.043 | 0.760 | +/- | 0.037 | 0.769 | +/- | 0.054 |
| Chemception_T2_F32 | 0.810 | +/- | 0.044 | 0.760 | +/- | 0.034 | 0.772 | +/- | 0.056 |
| Chemception_T2_F64 | 0.806 | +/- | 0.047 | 0.759 | +/- | 0.033 | 0.766 | +/- | 0.055 |
| **Chemception_T3_F16** | **0.815** | **+/-** | **0.044** | **0.768** | **+/-** | **0.037** | **0.773** | **+/-** | **0.058** |
| Chemception_T3_F32 | 0.814 | +/- | 0.045 | 0.763 | +/- | 0.034 | 0.771 | +/- | 0.055 |
| Chemception_T3_F64 | 0.765 | +/- | 0.046 | 0.733 | +/- | 0.039 | 0.739 | +/- | 0.052 |

We also note the relatively similar performance of Chemception T1 to T5. These architectures span between 21 and 69 layers but all have the same AUC to the second decimal place. As discussed in the original ResNet paper,[2] this is likely due to the effect that residual links have, where extraneous layers learn the identity mapping and become a computational cost but not a cost to accuracy, and it possibly suggests that a well-designed architecture at the high-level can be reasonably robust to "suboptimal" choices in the minutiae of architecture design.

### 3.5    Possible Further Optimizations

While we have performed and reported a series of Chemception architecture optimization, we should emphasize that there may be still additional room for improvement. Specifically, we did not vary the number of inception blocks across the 3 segments. As each inception block views a distinct spatial level, and it is only the reduction block that significantly changes its spatial level, one might expect that more features could be learned at some levels relative to others. If this is the case, then keeping the convolutional filters uniform across layers and segments may also not provide optimal expressive power at each spatial level. In addition, we have retained several design choices of the Inception module from Inception-ResNet v2, such as the design of the blocks, the number of inception branches, the ratio of filters from one layer to the next within the inception block, the choice of pooling layers, etc. Further fine tuning in the neural network architecture at this level may lead to additional improvement. Lastly, we also used a standardized training protocol using RMSProp, followed by SGD fine-tuning, but it is known that careful hyperparameter optimization of SGD learning rate have demonstrated superior results as evidenced from the winning entries of various ImageNet competitions.[52] Despite these additional avenues for possible improvements, we emphasize that the focus of this work is to determine the plausibility of using deep neural networks like Chemception as a "machine intelligence" tool; more aggressive architecture and hyperparameter tuning to squeeze out the last drop of the performance metric is thus not within the scope of this paper.

### 3.6　Tox21 Results

Using the best results obtained from the Chemception series of network architectures, we now compare our results with expert-developed QSAR/QSPR models reported in the literature. In particular, the most comparable results which controls for other factors, particularly data, can be obtained from the MoleculeNet paper released by Pande and co-workers.[39] In that work, they tested a deep neural network based on a multi-task multi-layer perceptron (MT-MLP) architecture, using engineered features in the form of ECFP fingerprints as inputs, and we have included their results for random data splitting for comparison. The single-task Chemception that we have developed should best be compared against a single-task MLP DNN which was not reported. The next most appropriate comparison thus falls to a multi-task MLP DNN, which is expected to perform better due to multi-task learning.[9, 11-12] As shown in **Figure 6 and Table S2**, despite the lack of engineered features and chemical knowledge, the single-task Chemception achieves a validation/test AUC of 0.768/0.773, against a multi-task MLP DNN with a validation/test AUC of 0.777/0.799 respectively, which is almost as good as a multi-task DNN that uses engineered features.

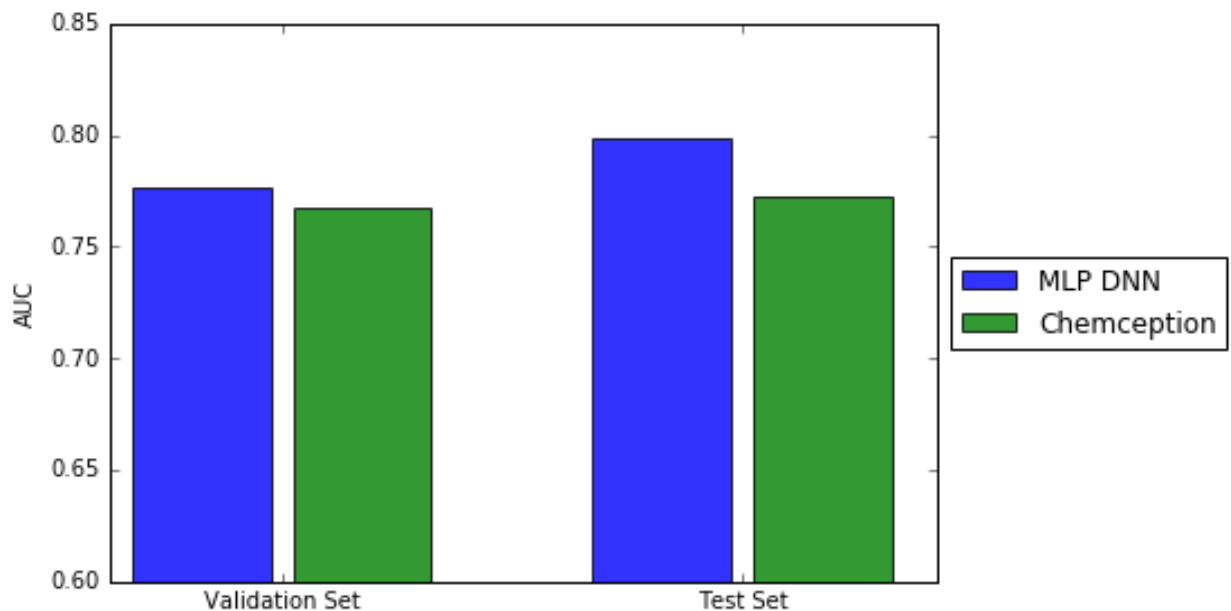

**Figure 6:** Comparison of Chemception performance to the reported results of MoleculeNet benchmark data on the Tox21 dataset.

### 3.7　HIV and FreeSolv Results

Having demonstrated that Chemception, trained with minimal prior chemistry knowledge can almost match the performance of comparable deep neural network models, we now address the question of whether Chemception would be able to repeat its performance on other chemical properties. Using the same training protocol, we predicted both HIV activity, a biochemical property, and free energy of solvation, a physical property, using both the baseline (T1) and the optimized (T3_F16) Chemception architecture. These measurements can be considered as separate "types" of chemical properties from the Tox21 measurements, and thus may require a different set of features to be learned if the network is going to achieve a reasonable level of performance.

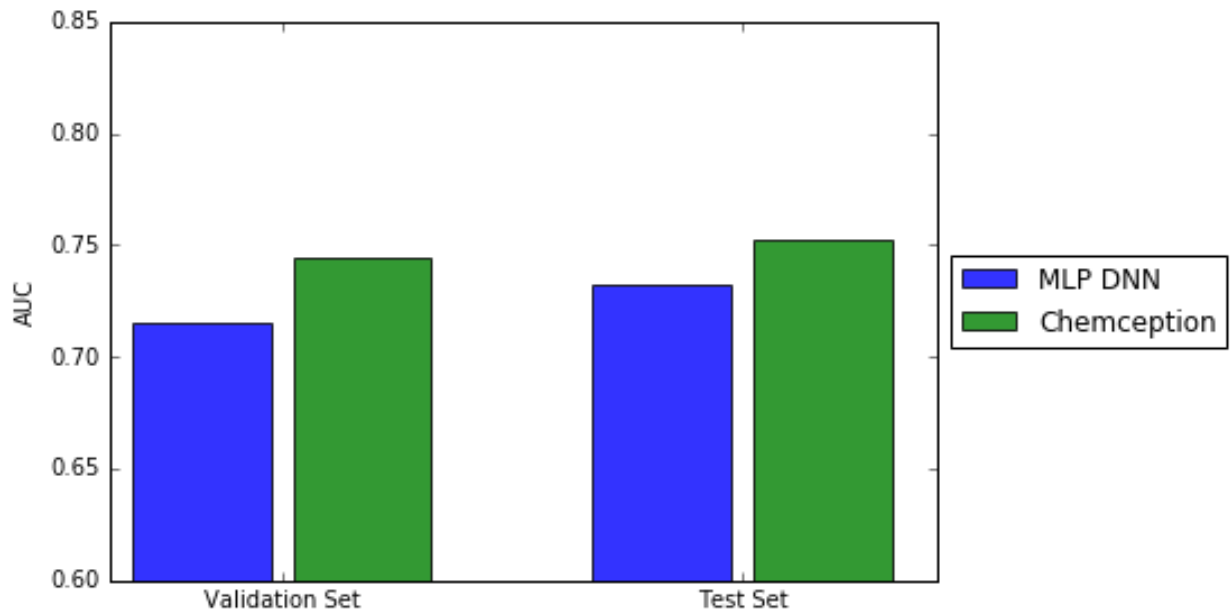

**Figure 7:** Comparison of Chemception performance to the reported results of MoleculeNet benchmark data on the HIV dataset.

Similar to the Tox21 dataset, the HIV activity prediction is a binary classification problem, but it has substantially more data (~40,000) than the Tox21 dataset (~8,000). In addition, unlike Tox21, there is only a single task to predict, and without multitask learning, it would thus provide a more comparable evaluation of Chemception performance against contemporary models. Interestingly, unlike the Tox21 dataset, for HIV, the baseline and finalized model performed similarly in terms of validation AUC, although there is a slight improvement in the test AUC from 0.744 to 0.752. Our results summarized in **Figure 7 and Table S3** indicates that the single-task Chemception with a validation/test AUC of 0.744/0.752 outperforms its counterpart single-task MLP DNN (validation/test AUC of 0.742/0.715). A plausible factor that explains Chemception improved performance on the HIV dataset relative to Tox21 is the increased size of the data provided. As Chemception is provided with minimal chemical knowledge and engineered features, it has to develop its own representation, and more data would thus make it easier for it to do so. In addition, we note that the HIV dataset has one task, and thus does not benefit from multi-task learning, which we earlier suggested may be responsible for the slight underperformance of Chemception in the Tox21 dataset against a multi-task MLP DNN.

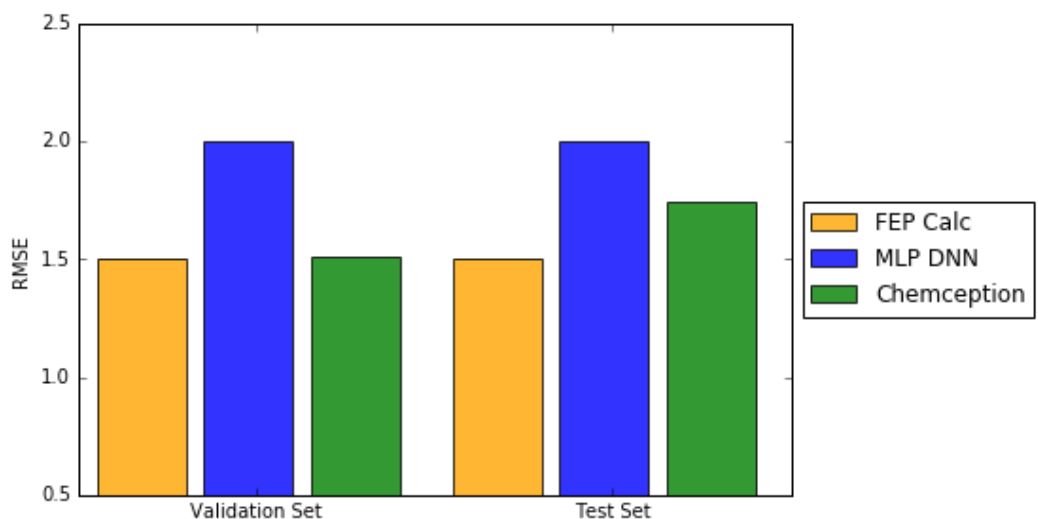

**Figure 8:** Comparison of Chemception performance to the reported results of MoleculeNet benchmark data on the FreeSolv dataset.

The last dataset that we trained Chemception on is to predict is solvation free energies. As summarized in **Figure 8 and Table S4**, the baseline Chemception model achieved a solvation energy RMSE error of 1.67 kcal/mol on the validation set and 1.84 kcal/mol on the test set. Similar to the trends observed in the Tox21 dataset, the finalized Chemception model achieved better performance with RMSE error of 1.51 kcal/mol and 1.75 kcal/mol for validation and test set respectively. When compared against its counterpart MLP DNN, which achieved a test RMSE of ~2.0 kcal/mol when 90% of the original dataset was used for training, Chemception provides a clear outperformance. Furthermore, unlike toxicity and activity modeling, where the use of physics-based model is limited, solvation free energies can be computed from free energy calculations from molecular dynamics simulations, and as such their computed values provide an additional comparison of physics-based models to QSPR-based models. Based on the FreeSolv dataset, it can be computed that the RMSE error for simulation-based method is ~1.5 kcal/mol. With its current performance, Chemception (~1.75 kcal/mol) has yet to attain the accuracy of simulation-based methods, but the results nevertheless is impressive because it demonstrates that even with an extremely limited quantity of data (~600 compounds), and minimal prior chemical knowledge, Chemception can attain an accuracy that is approaching state-of-the-art physics-based models. Similar efforts in training deep neural networks to predict QM-calculated energies have demonstrated that as the size of the training data increases, so does the model's accuracy.[22] This suggests that a larger dataset on solvation free energies is likely to improve Chemception performance, to reach parity with physics-based models.

## 4. Conclusion

In conclusion, we have developed a novel deep convolutional neural network, Chemception, for the prediction of chemical properties using only image data of 2D drawings of molecules. Without providing *any* further explicit chemistry knowledge, such as in the form of engineered features like molecular descriptors or fingerprints, we have demonstrated that the Chemception architecture can serve as a general-purpose neural network for learning a range of distinct properties, including physiological (toxicity), biochemical (activity) and physical (free energy of solvation) properties of molecules, while using a modest training database ranging from only ~600 to ~40,000 compounds. In addition, the general accuracy of Chemception across 3 tasks matches MLP deep neural networks trained on engineered features, such as ECFP fingerprints. For 2 out of the 3 properties, Chemception outperforms its deep neural network counterparts, and it achieved a validation/test AUC of 0.744/0.752 for HIV activity prediction and a validation/test RMSE of 1.51/1.74 kcal/mol for free energy of solvation prediction, which is close to the accuracy of physics-based simulation methods (RMSE ~1.5 kcal/mol). For the toxicity predictions on the Tox21 dataset, Chemception (validation/test AUC of 0.768/0.773) trails slightly in performance against multi-task DNN trained on ECFP fingerprints (validation/test AUC of 0.777/0.799), possibly due to the multi-task learning benefits employed in contemporary models. Inspired by the success of using deep learning as a "machine intelligence" tool in modern computer vision research, we have demonstrated the plausibility of using deep neural networks "machine intelligence" to assist the human-driven feature engineering step in the scientific discovery process. Furthermore, given that deep neural networks can process data at a much higher velocity and more consistently than humans can, coupled with the exponential growth of chemical data on which to train these networks, we anticipate that deep neural networks will be a valuable tool in the future of "machine intelligence" assisted computational chemistry research. Lastly, as we have taken a domain agnostic approach in developing Chemception, minimizing the amount of chemistry knowledge used in training it, we anticipate that the general approach used will also be transferable to other problems, which will be particularly useful in situations where there is limited understanding and feature engineering.

**Acknowledgement**

This research was funded by the Pacific Northwest National Laboratory (PNNL), Laboratory Directed Research Development (LDRD) Program for the Linus Pauling Distinguished Postdoctoral Fellowship and the Deep Learning for Scientific Discovery Initiative. Additional funding was provided through DOE ASCR grant 69696. Computing resources were provided on the XSEDE supercomputing grant CHE170015.

**References**

1. Krizhevsky, A.; Sutskever, I.; Hinton, G. E. ImageNet Classification with Deep Convolutional Neural Networks *Advances in Neural Information Processing Systems* **2012**.
2. He, K.; Zhang, X.; Ren, S.; Sun, J. Delving Deep into Rectifiers: Surpassing Human-Level Performance on ImageNet Classificatio. *arXiv:1502.01852* **2015**.
3. Ioffe, S.; Szegedy, C. Batch Normalization: Accelerating Deep Network Training by Reducing Internal Covariate Shift. *arXiv:1502.03167* **2015**.
4. Baldi, P.; Sadowski, P.; Whiteson, D. Searching for exotic particles in high-energy physics with deep learning. *Nat. Commun.* **2014,** *5*, 4308.
5. Charles Siegel; Jeff Daily; Vishnu, A. Adaptive Neuron Apoptosis for Accelerating Deep Learning on Large Scale Systems. *arXiv:1610.00790* **2016**.
6. Nikhil Mukund; Sheelu Abraham; Shivaraj Kandhasamy; Sanjit Mitra; Philip, N. S. Transient Classification in LIGO data using Difference Boosting Neural Network. *arXiv:1609.07259* **2017**.
7. Chicco, D.; Sadowski, P.; Baldi, P. Deep autoencoder neural networks for gene ontology annotation predictions. *Proc. of the 5th ACM Conf. on Bioinf. Comput. Biol. and Health Inf.* **2014**, 533-540.
8. Weihua Guo; You Xu; Feng, X. DeepMetabolism: A Deep Learning System to Predict Phenotype from Genome Sequencing. *arXiv:1705.03094* **2017**.
9. Dahl, G. E.; Jaitly, N.; Salakhutdinov, R. Multi-task Neural Networks for QSAR Predictions *arXiv:1406.1231* **2014**.
10. Ma, J.; Sheridan, R. P.; Liaw, A.; Dahl, G. E.; Svetnik, V. Deep neural nets as a method for quantitative structure-activity relationships. *J. Chem. Inf. Model.* **2015,** *55* (2), 263-74.
11. Ramsundar, B.; Kearnes, S.; Riley, P.; Webster, D.; Konerding, D.; Pande, V. Massively Multitask Networks for Drug Discovery. *arXiv:1502.02072* **2015**.
12. Unterthiner, T.; Mayr, A.; Klambauer, G.; Steijaert, M.; Ceulemans, H.; Wegner, J.; Hochreiter, S. Multi-Task Deep Networks for Drug Target Prediction. *Conference Neural Information Processing Systems Foundation (NIPS 2014)* **2014**.
13. Mayr, A.; Klambauer, G.; Unterthiner, T.; Hochreiter, S. DeepTox: Toxicity Prediction using Deep Learning. *Front. Env. Sci.* **2016,** *3*, 1-15.
14. Xu, Y.; Dai, Z.; Chen, F.; Gao, S.; Pei, J.; Lai, L. Deep Learning for Drug-Induced Liver Injury. *J. Chem. Inf. Model.* **2015,** *55* (10), 2085-93.
15. Hughes, T. B.; Dang, N. L.; Miller, G. P.; Swamidass, S. J. Modeling Reactivity to Biological Macromolecules with a Deep Multitask Network. *ACS Cent. Sci.* **2016,** *2* (8), 529-37.
16. Hughes, T. B.; Miller, G. P.; Swamidass, S. J. Modeling Epoxidation of Drug-like Molecules with a Deep Machine Learning Network. *ACS Cent. Sci.* **2015,** *1* (4), 168-80.
17. Hughes, T. B.; Miller, G. P.; Swamidass, S. J. Site of Reactivity Models Predict Molecular Reactivity of Diverse Chemicals with Glutathione. *Chem. Res. Toxicol.* **2015,** *28* (4), 797-809.
18. Lusci, A.; Pollastri, G.; Baldi, P. Deep architectures and deep learning in chemoinformatics: the prediction of aqueous solubility for drug-like molecules. *J. Chem. Inf. Model.* **2013,** *53* (7), 1563-75.
19. Kearnes, S.; Goldman, B.; Pande, V. Modeling Industrial ADMET Data with Multitask Networks. *arXiv:1606.08793* **2016**.

20. Wallach, I.; Dzamba, M.; Heifets, A. AtomNet: A Deep Convolutional Neural Network for Bioactivity Prediction in Structure-based Drug Discovery. *arXiv:1510.02855* **2016**.
21. Montavon, G.; Rupp, M.; Gobre, V.; Vazquez-Mayagoitia, A.; Hansen, K.; Tkatchenko, A.; Müller, K.-R.; Anatole von Lilienfeld, O. Machine learning of molecular electronic properties in chemical compound space. *New J. Phys.* **2013,** *15* (9), 095003.
22. Smith, J. S.; Isayev, O.; Roitberg, A. E. ANI-1: an extensible neural network potential with DFT accuracy at force field computational cost. *Chem Sci* **2017,** *8* (4), 3192-3203.
23. Schutt, K. T.; Arbabzadah, F.; Chmiela, S.; Muller, K. R.; Tkatchenko, A. Quantum-chemical insights from deep tensor neural networks. *Nat Commun* **2017,** *8*, 13890.
24. Goh, G. B.; Hodas, N. O.; Vishnu, A. Deep learning for computational chemistry. *Journal of computational chemistry* **2017,** *38* (16), 1291-1307.
25. Gawehn, E.; Hiss, J. A.; Schneider, G. Deep Learning in Drug Discovery. *Mol. Inf.* **2016,** *35* (1), 3-14.
26. Lowe, D. G. Object recognition from local scale-invariant features. *Proceedings of the International Conference on Computer Vision* **1999**, 1150–1157.
27. Silver, D.; Huang, A.; Maddison, C. J.; Guez, A.; Sifre, L.; van den Driessche, G.; Schrittwieser, J.; Antonoglou, I.; Panneershelvam, V.; Lanctot, M.; Dieleman, S.; Grewe, D.; Nham, J.; Kalchbrenner, N.; Sutskever, I.; Lillicrap, T.; Leach, M.; Kavukcuoglu, K.; Graepel, T.; Hassabis, D. Mastering the game of Go with deep neural networks and tree search. *Nature* **2016,** *529* (7587), 484-+.
28. Wu Y, e. a. Google's Neural Machine Translation System: Bridging the Gap between Human and Machine Translation. *arXiv:1609.08144* **2016**.
29. Szegedy, C.; Liu, W.; Jia, Y.; Sermanet, P.; Reed, S.; Anguelov, D.; Erhan, D.; Vanhoucke, V.; Rabinovich, A. Going deeper with convolutions. *arXiv:1409.4842* **2014**.
30. He, K.; Zhang, X.; Ren, S.; Sun, J. Deep Residual Learning for Image Recognition. *arXiv:1512.03385* **2015**.
31. Wiener, H. Structural Determination of Paraffin Boiling Points. *J. Am. Chem. Soc.* **1947,** *69* (1), 17-20.
32. Platt, J. R. Influence of Neighbor Bonds on Additive Bond Properties in Paraffins. *J. Chem. Phys.* **1947,** *15* (6), 419-420.
33. O'Boyle, N. M.; Banck, M.; James, C. A.; Morley, C.; Vandermeersch, T.; Hutchison, G. R. Open Babel: An open chemical toolbox. *J Cheminform* **2011,** *3*, 33.
34. O'Boyle, N. M.; Morley, C.; Hutchison, G. R. Pybel: a Python wrapper for the OpenBabel cheminformatics toolkit. *Chem Cent J* **2008,** *2*, 5.
35. RDKit, O.-S. C. http://www.rdkit.org.
36. O'Boyle, N. M.; Hutchison, G. R. Cinfony--combining Open Source cheminformatics toolkits behind a common interface. *Chem Cent J* **2008,** *2*, 24.
37. Weininger, D. Smiles, a Chemical Language and Information-System .1. Introduction to Methodology and Encoding Rules. *J Chem Inf Comp Sci* **1988,** *28* (1), 31-36.
38. Weininger, D.; Weininger, A.; Weininger, J. L. Smiles .2. Algorithm for Generation of Unique Smiles Notation. *J Chem Inf Comp Sci* **1989,** *29* (2), 97-101.
39. Zhenqin Wu; Bharath Ramsundar; Evan N. Feinberg; Joseph Gomes; Caleb Geniesse; Aneesh S. Pappu; Karl Leswing; Pande, V. MoleculeNet: A Benchmark for Molecular Machine Learning. *arXiv:1703.00564* **2017**.
40. Huang, R.; Sakamuru, S.; Martin, M. T.; Reif, D. M.; Judson, R. S.; Houck, K. A.; Casey, W.; Hsieh, J. H.; Shockley, K. R.; Ceger, P.; Fostel, J.; Witt, K. L.; Tong, W.; Rotroff, D. M.;

Zhao, T.; Shinn, P.; Simeonov, A.; Dix, D. J.; Austin, C. P.; Kavlock, R. J.; Tice, R. R.; Xia, M. Profiling of the Tox21 10K compound library for agonists and antagonists of the estrogen receptor alpha signaling pathway. *Sci. Rep.* **2014,** *4*, 5664.
41. Mobley, D. L.; Guthrie, J. P. FreeSolv: a database of experimental and calculated hydration free energies, with input files. *Journal of computer-aided molecular design* **2014,** *28* (7), 711-20.
42. Data, A. A. S. https://wiki.nci.nih.gov/display/NCIDTPdata/AIDS+Antiviral+Screen+Data.
43. Bengio, Y.; Courville, A.; Vincent, P. Representation Learning. *IEEE Trans. Pattern Anal. Mach. Intell.* **2013,** *35* (8), 1798-1827.
44. Schmidhuber, J. Deep learning in neural networks: an overview. *Neural Netw.* **2015,** *61*, 85-117.
45. Arel, I.; Rose, D. C.; Karnowski, T. P. Deep Machine Learning-A New Frontier in Artificial Intelligence Research. *IEEE Comput. Intell. M.* **2010,** *5* (4), 13-18.
46. Glorot, X.; Bordes, A.; Bengio, Y. Deep Sparse Rectifier Neural Networks. *Proc. of the 14th Int. Conf. on Artificial Intelligence and Statistics (AISTATS)* **2011**.
47. Szegedy, C.; Ioffe, S.; Vanhoucke, V.; Alemi, A. Inception-v4, Inception-ResNet and the Impact of Residual Connections on Learning. *arXiv:1602.07261* **2016**.
48. Martín Abadi, e. a. TensorFlow: A system for large-scale machine learning. *arXiv:1605.08695* **2016**.
49. Sharan Chetlur; Cliff Woolley; Philippe Vandermersch; Jonathan Cohen; John Tran; Bryan Catanzaro; Shelhamer, E. cuDNN: Efficient Primitives for Deep Learning. *arXiv:1410.0759* **2014**.
50. Chollet, F. Keras. *https://github.com/fchollet/keras* **2015**.
51. Tieleman, T.; Hinton, G. RMSProp: Divide the gradient by a running average of its recent magnitude. *COURSERA: Neural Networks for Machine Learning* **2012**.
52. Russakovsky, O.; Deng, J.; Su, H.; Krause, J.; Satheesh, S.; Ma, S.; Huang, Z.; Karpathy, A.; Khosla, A.; Bernstein, M.; Berg, A. C.; Fei-Fei, L. ImageNet Large Scale Visual Recognition Challenge. *IJCV* **2015,** *115* (3), 211-252.
53. Ribeiro, M. T.; Singh, S.; Guestrin, C. "Why Should I Trust You?" Explaining the Predictions of Any Classifie. *arXiv:1602.04938* **2016**.
54. Wu., G. http://data-informed.com/why-more-data-and-simple-algorithms-beat-complex-analytics-models/**2013**.
55. Schnoebelen, T. https://www.crowdflower.com/more-data-beats-better-algorithms/**2016**.